\RequirePackage{fix-cm}
\documentclass{svjour3}                     
\smartqed  
\usepackage{graphicx}
\usepackage{amsmath}
\usepackage{color,soul}
\begin{document}

\title{Rigid vs compliant contact: An experimental study on biped walking}


\author{Majid Khadiv         \and
        S. Ali A. Moosavian \and
        Aghil Yousefi-Koma \and
         Majid Sadedel \and
         Abbas Ehsani-Seresht \and
         Saeed Mansouri
}


\institute{Majid Khadiv \at
              Movement Generation and Control Group, Max Planck Institute for Intelligent Systems, Tuebingen, Germany. \\
              \email{mkhadiv@tuebingen.mpg.de}          
           \and
           S. Ali A. Moosvian \at
              Department of Mechanical Engineering K. N. Toosi. University of Technology, Tehran, Iran. \\
              \email{moosavian@kntu.ac.ir}
            \and
           Aghil Yousefi-Koma \at
              School of Mechanical Engineering, College of Engineering, University of Tehran, Tehran, Iran.\\
              \email{aykoma@ut.ac.ir}
            \and
           Majid Sadedel \at
            Department of mechanical engineering, Tarbiat Modares University, Tehran, Iran.\\
            \email{majid.sadedel@modares.ac.ir}
            \and
           Abbas Ehsani-Seresht \at
              Department of Mechanical Engineering, Hakim Sabzevari University, Sabzevar, Iran.\\
              \email{a.ehsaniseresht@hsu.ac.ir}  
              \and
           Saeed Mansouri \at
              Department of Mechanical Engineering, Sharif University of Technology, Tehran, Iran.\\
              \email{s\underline{ }mansouri@mech.sharif.edu} 
}

\date{Received: date / Accepted: date}

\maketitle

\begin{abstract}
Contact modeling plays a central role in motion planning, simulation and control of legged robots, as legged locomotion is realized through contact. The two prevailing approaches to model the contact consider rigid and compliant premise at interaction ports. Contrary to the dynamics model of legged systems with rigid contact (without impact) which is straightforward to develop, there is no consensus among researchers to employ a standard compliant contact model. Our main goal in this paper is to study the dynamics model structure of bipedal walking systems with rigid contact and a \textit{novel} compliant contact model, and to present experimental validation of both models. For the model with rigid contact, after developing the model of the articulated bodies in flight phase without any contact with environment, we apply the holonomic constraints at contact points and develop a constrained dynamics model of the robot in both single and double support phases. For the model with compliant contact, we propose a novel nonlinear contact model and simulate motion of the robot using this model.  In order to show the performance of the developed  models, 
 we compare obtained results from these models to the empirical measurements from bipedal walking of the human-size  humanoid  robot  SURENA  III,  which  has  been  designed  and  fabricated  at  CAST,  University  of  Tehran.  This 
analysis shows the merit of both models in estimating dynamic behavior of the robot walking on a semi-rigid surface. The model with rigid contact, which is less 
complex and  independent of the physical properties of the contacting bodies, can  be  employed  for  model-based  motion  optimization,  analysis  as  well  as  control,  while  the  model  with compliant contact and more complexity is suitable for more realistic simulation scenarios. 
\end{abstract}

\section{Introduction}
\label{Introduction}
Due to the complex structure of humanoid robots and their nonlinear and hybrid dynamics, developing a 
tool for simulating these sophisticated machines is significantly important. This tool may be exploited in the hardware 
design procedure, optimization-based motion planning and simulation, and model-based controller design process of 
humanoid robots.  

Regarding hardware selection in the design procedure of  humanoid robots, we need a comprehensive dynamics model including motors and drive system dynamics, sensors and multibody dynamics, and contact mechanics \cite{buschmann2010simulation,buschmann2009humanoid}.
In this stage, there is no constraint on the computation load and cost. Therefore, a thorough dynamics model or simulation environment in this 
stage will lead to an appropriate components choice for the robot \cite{komoda2017energy,mazumdar2017parallel}. 

To generate walking patterns for humanoid robots in real-time, methods based on an abstract model of the robot dynamics \cite{kajita20013d} have been successfully deployed  \cite{englsberger2015three,faraji2014robust,feng2016robust,herdt2010online,khadiv2016step,khadiv2016stepping}. Furthermore, for more complicated tasks, the centroidal momentum dynamics \cite{orin2013centroidal} has been shown to be very effective \cite{carpentier2016A,dai2016planning,herzog2016structured,ponton2016convex,tassa2012synthesis}.  However, in order to optimize the motion in terms of actuation or energetics, we need a full dynamics model of the robot \cite{kim2014numerical,lim2014gait,tlalolini2010design}. Since the full dynamics of a humanoid robot is high-dimensional, we are not able to employ them to generate motions in real-time based on the state-of-the-art computational power. However, as the technology progresses and the mathematical tools become mature, the use of full dynamics to generate efficient motions in real-time becomes relevant.  

In order to map generated walking patterns to the full body of a humanoid robot, various whole body controllers have been proposed \cite{herzog2016momentum,hopkins2016optimization,koolen2016design}. These controllers use the full dynamics of the robot to compute instantaneous joint torques at each control cycle consistent with the desired tasks and physical constraints. Having a precise dynamics model of the robot, which can be evaluated very rapidly  results in a high frequency update of the control inputs (for instance 1 kHz in \cite{herzog2016momentum}).

In  order  to  derive  a  complete dynamics model, we need to take into account the effects of various components that take part in the motion of the robot. The major  effects  for  a legged robot  are  the  articulated rigid body dynamics and the  robot-environment  contact  mechanics \cite{buschmann2010simulation}. Although there is a consensus among researchers for modeling rigid body dynamics of legged robots \cite{featherstone2014rigid}, contact modeling is more debatable and there is an ongoing research in this area \cite{wieber2016modeling}.

Humanoid robots interact with  the environment through their feet (and for more complex tasks through their hands \cite{carpentier2016A,nikolic2017dynamic,ponton2016convex}), and this contact should be modeled properly \cite{lopes2016superellipsoid,brown20183d}. In this notion, researchers typically adopt two approaches \cite{wieber2016modeling}. In the first 
approach, the contact between the feet and ground surface is considered to be 
rigid. In this method, it is assumed that there is no deflection between the feet and ground surface, and 
complementary condition is considered at each contact point. Then, consistent with constraints, forces and moments 
are considered and mapped to the joint space of the robot. Since contact points vary during different phases, the 
dynamics  model  is  different  during  various  phases  of  the  motion.  Second  approach  uses  compliant  elements  to 
model  the  contact.  In  this  method,  springs  and  dampers  at  contact  points  are  assumed.  Besides  some 
researchers who adopted linear springs and dampers \cite{buschmann2010simulation,herzog2015trajectory,peasgood2007stabilization,pratt2012capturability,righetti2013optimal}, others exploited nonlinear elements to achieve a model that is 
more compatible with the reality \cite{jackson2016development,mclean2003development,millard2009multi,park2001reflex,wojtyra2003multibody}. In this approach which is known as penalty method, the unilaterality conditions are added to springs and dampers. 

Contrary to the dynamics model of legged systems with rigid contact (without impact) which is straightforward to compute, there is no consensus among researchers to employ a standard compliant contact model. Furthermore, to the best of our knowledge, none of previous works have done a thorough analysis on the attributes of each model and their differences. Consequently, the contribution of this work is twofold. First, we propose a novel compliant contact model which satisfies the potential requirements. Second, we present experimental validation of both models with rigid contact and the proposed compliant contact, and discuss the differences between these models and their usages. It is noteworthy that our goal in this paper is not to compare the precision of models to show which one is better, as the merit of the models are highly dependant on their application.\\
The procedure of developing models with rigid and compliant contact is as follows:  
\begin{itemize}
  \item  \textit{Model with rigid contact (without impact).} In  this  approach,  we first develop the robot dynamics  without  interaction  with  the environment  (flight  phase).  Then,  in  order  to  take  into  account  the  interaction  between  the  feet  and ground surface, we enforce a rigid model of contact at interaction points. In this approach, we consider holonomic constraints at each interaction point, and using the constraint relaxation method \cite{baruh1999analytical}, we replace each constraint with an  unknown  force/moment.  Finally,  we find the  solution  of  the  inverse  dynamics problem   to  compute  the  joint torques and interaction forces/moments for a given motion. This approach is relatively simple and employs 
some  assumptions,  i.  e.,  the  motion  is  impactless, the  interaction  forces/moments  are 
feasible, the motion phase is given, etc. Noteworthy is that if  modeling  the  impact  is  essential  as  in  the  passive walkers  case, we need to go through modeling this phenomenon in our model. Although we may simply take into account the effect of impact in the rigid contact model with an instantaneous change in the velocity after the impact, for a better representation of impact, we also need a model of impulsive forces. However, adding the  model  of impulsive  forces  for  computing  the  post-impact  velocity  increases  significantly  the  complexity  of  the  model \cite{jia2013multiple}.\\

  \item \textit{Model with compliant contact.} In  this  approach,  we model the  interaction   employing  a  compliant  contact  model.  We propose a novel nonlinear contact model which satisfies the requirements for a realistic simulation. Exploiting this novel 
contact model, and modeling the multibody of the robot in an available physics engine, we simulate the motion 
of the robot with a compliant contact model. Though the model with a compliant contact is more complicated compared to that with a rigid contact model, it uses fewer assumptions. For example, impacts are simulated, or knowledge about the motion phase  is not required in advance. As a result, this approach is more suitable for 
simulating motion of the robot in a more realistic scenario. However, the complexity of this approach and its dependence on prior knowledge of the contact properties limits its 
application in the motion optimization or the design of model-based controllers.
\end{itemize}

\begin{figure}
\centering
\includegraphics[clip,trim=6cm 19.35cm 7cm 2.4cm,width=8cm]{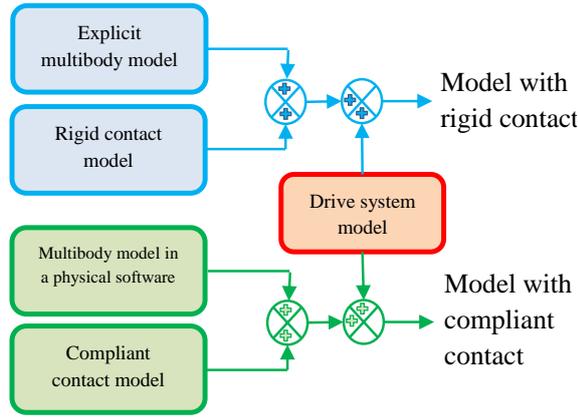}
\caption{The procedure of developing model with rigid and compliant contact in this study}
\vspace{-1.5em}
\label{block_diagram}
\end{figure}

In order to improve the models’ precision, we add the identified drive dynamics to the computed torques of both models 
(see  Fig.  \ref{block_diagram} for a big picture of the whole procedure).  Then,  obtained  results  from  the  two  methods  are  compared  and  their  differences,  as well as  their advantages/disadvantages are discussed.  
The rest of this paper is structured as follows: In Sect.~\ref{rigid}, we present the multibody dynamics modeling with rigid contact. We dedicate Sect. ~\ref{compliant} to modeling of the robot in a physics engine with compliant  interaction.  In  Sect.~\ref{simulation},  we present and discuss obtained  simulation  and  experimental results from a 3D humanoid bipedal walking. Finally, in Sect.~\ref{conclusion} we conclude the findings.

\section{Model with rigid contact}
\label{rigid}

\begin{figure}
\centering
\includegraphics[clip,trim=7cm 17cm 7cm 3cm,width=6cm]{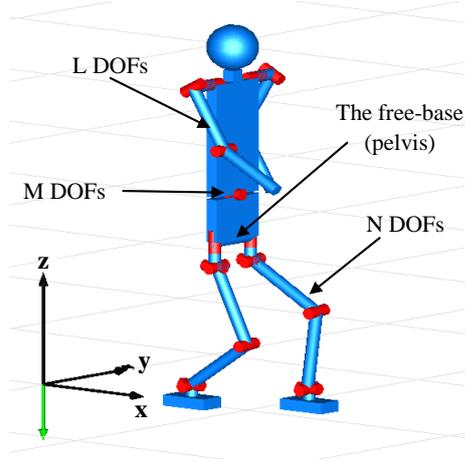}
\caption{Model of a humanoid robot (SURENA III humanoid robot) walking on a flat surface}
\vspace{-1.5em}
\label{robot}
\end{figure}

The set of generalized coordinates which describes the motion of a humanoid robot is considered as
\begin{align}
\label{gen_coord}
q=\begin{bmatrix} q_p &  q_{rl} & q_{ll}  &  q_{tr}&  q_{rh} & q_{lh}    	\end{bmatrix} ^T
\end{align}

where $q_p \in \Re^{6}$ is a vector which describes motion of  the pelvis  (the free-floating generalized coordinates)  with respect to the inertial coordinate system. $q_{rl}, q_{ll} \in \Re^{N}$  are the right and left leg joint  vectors, while $q_{rh}, q_{lh} \in \Re^{L}$ are the vectors which describe arm joint angles;  $q_{tr} \in \Re^{M}$ is the vector of the upper-body joints (trunk and waist joints, see Fig. \ref{robot}).

The set of equations of motion for the robot with no interaction with the environment can be stated as
\begin{align}
\label{dynamics}
M(q)\ddot q+V(q,\dot q)+G(q)=Q
\end{align}

In this equation, $M \in \Re^{(6+M+2N+2L)\times(6+M+2N+2L)}$ is the generalized inertia matrix, $V \in \Re^{(6+M+2N+2L)}$ groups together the Coriolis as well as centrifugal effects, and $G \in \Re^{(6+M+2N+2L)}$ contains the gravity terms. Moreover, $Q \in \Re^{(6+M+2N+2L)}$ is the vector which specifies generalized forces acting on the robot. This term varies during different phases of the motion. In the case that the robot does not interact with the environment (flight phase), $Q$ is:
\begin{align}
\label{gen_force}
Q=B \tau
\end{align}

where $\tau \in \Re^{(M+2N+2L)}$ is a vector including joint torques and can be stated as
\begin{align}
\label{tau}
\tau=\begin{bmatrix}  \tau_{rl} & \tau_{ll}  &  \tau_{tr}&  \tau_{rh} & \tau_{lh}    	\end{bmatrix} ^T
\end{align}

Moreover, $B$ is a constant matrix which projects joint torques to the space of generalized coordinates. Since we used the relative joint angles as generalized coordinates, matrix $B$ is
\begin{align}
\label{B}
B=\begin{bmatrix}  0_{(M+2N+2L)\times 6} & I_{(M+2N+2L)\times (M+2N+2L)}    	\end{bmatrix} ^T
\end{align}

Now, in order to obtain the constrained dynamics model in different phases of motion with various interactions, we specify the constraints in each phase in the rest of this section. It should be noted that here we just investigate flat-feet walking phases in this paper. We investigated other gait phases such as toe-off and heel-contact in our earlier works \cite{ezati2015effects,sadedel2016adding,sadedel2016investigation}.

\subsection{Single Support Phase (SSP)}

The Single Support Phase (SSP) is the phase when one of the feet is fixed to the ground surface and the other moves on a desired trajectory. In the case that the stance foot does not slip or rotate around its edges, it can be assumed that this foot is fixed to the ground. As long as the ZMP is within the support polygon during motion and no-slip constraints are satisfied, this assumption is valid. In fact, we consider the rigid contact model at the contact points which means that the contact points are stationary. The holonomic constraints that are used for the stance foot may be specified as
\begin{align}
\label{hol}
\begin{bmatrix}  X_s-C \\ \theta_0    	\end{bmatrix} ^T = const.
\end{align}

In this equation, $X_s$ represents the position of an arbitrary point on the stance foot and $\theta_0$ is the orientation of the stance foot, while $C$ is a constant vector. Using constraint relaxation method \cite{baruh1999analytical}, consistent with these constraints, we exert unknown interaction forces and moments in the reference point of the foot in contact with the ground surface. Then, using the transpose of Jacobian of contact points, we map these interaction components to the space of generalized coordinates. This way, in the SSP, the generalized force vector can be stated as
\begin{align}
\label{gen_force_ssp}
Q=B \tau+J^T F
\end{align}

In this equation, $F \in \Re^{6}$ is the vector including interaction forces and moments acting on the stance foot reference point. Furthermore, $J \in \Re^{6\times(6+M+2N+2L)}$ represents the Jacobian of the stance foot reference point resolved in the inertial coordinate system.

In order to solve the inverse dynamics problem, we write down the obtained dynamics model in the following compact form
\begin{align}
\label{dynamics_ssp}
[M\ddot q+V+G]_{(6+M+2N+2L)}=\begin{bmatrix}  B & J^T \end{bmatrix}
\begin{bmatrix}  \tau \\ F \end{bmatrix}_{(6+M+2N+2L)}
\end{align}

Since in this equation $\begin{bmatrix}  B & J^T \end{bmatrix} \in \Re^{(6+M+2N+2L)\times(6+M+2N+2L)}$ is a square matrix, the unique inverse dynamics solution is ($\begin{bmatrix}  B & J^T \end{bmatrix}$ should be invertible)
\begin{align}
\label{dynamics_ssp_sol}
\begin{bmatrix}  \tau \\ F \end{bmatrix}=\begin{bmatrix}  B & J^T \end{bmatrix}^{-1}[M\ddot q+V+G]
\end{align}

\subsection{Double Support Phase (DSP)}

The Double Support Phase (DSP) is the phase when both feet are in contact with the ground surface. In this phase, we apply the holonomic constraints of Eq. (\ref{hol}) to both feet, hence interaction forces are exerted on both feet. The generalized force vector in this phase is
\begin{align}
\label{gen_force_dsp}
Q=B \tau+J_{lf}^T F_{lf}+J_{rf}^T F_{rf}
\end{align}

In this equation, indexes $rf$ and $lf$ specify right and left foot interaction points, respectively. By substituting Eq. (\ref{gen_force_dsp}) into Eq. ({\ref{dynamics}}), the constrained dynamics model in the DSP can be obtained
\begin{align}
\label{dynamics_dsp}
[M\ddot q+V+G]_{(6+M+2N+2L)}=\begin{bmatrix}  B & J_{lf}^T&J_{rf}^T \end{bmatrix}
 \begin{bmatrix}  \tau \\  F_{lf} \\ F_{rf} \end{bmatrix}_{(12+M+2N+2L)}
\end{align}

In this case, the matrix $\begin{bmatrix}  B & J_{lf}^T&J_{rf}^T \end{bmatrix} \in \Re^{(6+M+2N+2L)\times(12+M+2N+2L)}$ is not squared. The reason is that during the DSP, the legs constitute a closed kinematic chain. In fact, the actuation redundancy in this phase makes the set of equations in (\ref{dynamics_dsp}) under-determined. As a result, the inverse dynamics problem has an infinite number of solutions. By defining $C$ and $D$ as
\begin{align}
\label{c}
C=\begin{bmatrix}  B & J_{lf}^T&J_{rf}^T \end{bmatrix}\\
D=M\ddot q+V+G
\end{align}

the set of all solutions of the inverse dynamics problem becomes
\begin{align}
\label{dynamics_dsp_sol}
\begin{bmatrix}  \tau \\  F_{lf} \\ F_{rf} \end{bmatrix}=C^{\dagger}D+(I-C^{\dagger}C)k
\end{align}

where $[ ]^{\dagger}$ is the Moore-Penrose inverse (right pseudo-inverse), $k$ is an arbitrary constant vector and $I$ is the identity matrix. One of the interesting solutions of this problem is
\begin{align}
\label{dynamics_dsp_psol}
\begin{bmatrix}  \tau \\  F_{lf} \\ F_{rf} \end{bmatrix}=C^{\dagger}D=\begin{bmatrix}  B & J_{lf}^T&J_{rf}^T \end{bmatrix}^{\dagger} [M\ddot q+V+G]
\end{align}

This solution set yields the minimum quadratic norm of the joint torques and interaction forces and moments. 

It is worth to note that the unconstrained inverse dynamics solution in this section is obtained for a set of feasible trajectories. In fact, we assumed that feasible trajectories are planned and mapped to the joint space using inverse kinematics \cite{khadiv2015optimal}, and then Eqs. (\ref{dynamics_ssp_sol}) and (\ref{dynamics_dsp_psol}) are used to compute actuation torques and interaction forces for the SSP and DSP. As a result, these solutions are valid as long as the stance foot (feet) is stationary and does not slip or tip over. The feasibility constraints may be taken into account using inequality constraints inside a whole body controller to generate feasible torque commands \cite{righetti2013optimal,wensing2016improved,herzog2016momentum}. However, since our main goal in this paper is just to compare the models with rigid and compliant contacts, we solved the unconstrained problem for a given feasible motion.

\section{Model with compliant contact}
\label{compliant}
In this section, we aim at modeling multiple bodies of the robot in a physics engine and simulating its motion with a compliant contact model. To do this, the robot with a specified number of DOF is modeled with its joints and links and free base (see Fig. \ref{robot}). The free base resembles the unactuated DOFs. Therefore, in order to model the robot in a physics engine, we should define a body (floating base, e. g., pelvis) which has 6 DOF with respect to the inertial coordinate system. Then, we connect each limb with its joints and links to the free base to obtain the multibody model. 

In order to model the contact between the feet and environment in the physics engine, we employ a compliant contact model. This model exploits springs and dampers at contact points to replicate compliant unilateral contact in real situation. In the rest of this section, after reviewing available compliant contact models in the literature, we propose a nonlinear contact model and adopt it for modeling the robot-environment interaction in the physics engine.

\subsection{Available models in the literature}
\noindent
A vast number of models have been suggested by researchers for compliant contact modeling between the feet and the ground surface. For contact modeling in the normal direction to the interacting bodies, the authors of \cite{peasgood2007stabilization,buschmann2010simulation,herzog2015trajectory,righetti2013optimal,pratt2012capturability} employed a linear viscoelastic model. In their model, the normal contact force is stated as:
\begin{align}
\label{con_lin}
F_N=-k_z \, \delta z-c_z \, \delta \dot z
\end{align}

where $\delta z$ and $\delta \dot z$̇ are the deflection and rate of penetration, respectively. Also, $k_z$ and $c_z$ are the stiffness and damping ratio of the springs and dampers. In this model, the unilateral contact constraints are taken into account using a penalty function. Wojtyra \cite{wojtyra2003multibody} used the same model with nonlinear dampers, where the damping ratio is a function of the penetration depth, namely
\begin{align}
\label{con_non1}
  c_z=\begin{cases}
              c_{max} |\frac{3 \delta z^2}{h^2}-\frac{2 \delta z^3}{h^3}| \quad &, \quad \delta z \leq h\\
              c_{max} &, \quad \delta z > h
            \end{cases}
\end{align}

In this equation, $c_{max}$ and $h$ are constants. McLean et al. \cite{mclean2003development} proposed a linear spring and nonlinear damper to model the normal contact force as
\begin{align}
\label{con_non2}
F_N=-k_z \,\delta z-b_z \, |\delta z| \, \delta \dot z
\end{align}

Similarly, Jackson et al. \cite{jackson2016development} exploited a nonlinear viscoelastic model to model the foot-ground interaction by
\begin{align}
\label{con_non_jack}
F_N=-k_z \,\delta z \, (1+c_z \, \delta \dot z)
\end{align}

where $k_z$ and $c_z$ are constant. Park and Kwon \cite{park2001reflex} exploited a nonlinear spring and damper to simulate motion of a biped robot. In this model, the normal contact force is formulated as
\begin{align}
\label{con_non3}
F_N(\delta_z)=-\frac{3}{2} \, \alpha \, k_z(\delta z) \, |\delta z| \, \delta \dot z-k_z(\delta z) \, \delta z
\end{align}

where $\delta z$ specifies the penetration of the contact point below the contact surface, $ k_z(\delta_z)$ is the deformation-dependent stiffness, $\alpha$ is a constant which defines the relation between the coefficient of restitution and the impact velocity. Also, \cite{millard2009multi} proposed a more complex model, namely
\begin{align}
\label{con_non4}
F_N=-k_z \,\delta z \, (1+\frac{1-\epsilon}{\epsilon \, \delta \dot z_0} \delta \dot z)
\end{align}

In this model, $\epsilon$ is the coefficient of restitution and $\delta z$̇ is the initial speed of impact. 

Researches on horizontal contact forces modeling can be divided into two categories. In the first category, springs and dampers are adopted in the horizontal direction. In this method, the horizontal forces are independent of the normal force. In this notion, the authors of \cite{buschmann2010simulation,herzog2015trajectory,righetti2013optimal} used linear springs and dampers to model the contact in horizontal direction. 

Park and Kwon \cite{park2001reflex} employed a linear spring and nonlinear damper, that is
\begin{align}
\label{con_hor2}
F_s=-\frac{3}{2} \alpha \, k_x \, |\delta x| \, \delta \dot x-k_s\delta_s
\end{align}

However, because in such models the horizontal contact forces are independent of the normal force, the model is not consistent with reality. To remedy this, some research studies used a modified Coulomb model to implement contact in horizontal directions. In this notion, \cite{peasgood2007stabilization} proposed the following model:
\begin{align}
\label{con_hor3}
F_s=-sign(\delta \dot x) \; \mu F_N
\end{align}

where $\delta \dot x$̇ is the horizontal velocity of the contact point; $\mu$ is the friction coefficient and is assumed to be different for dynamic and static cases:
\begin{align}
\label{mu_sta_dyn}
  \mu=\begin{cases}
              0.8 \quad &, \quad \delta \dot x \leq 0.05 m/s\\
             0.2 &, \quad \delta \dot x >  0.05 m/s
            \end{cases}
\end{align}

Wojtyra \cite{wojtyra2003multibody} used a different model:
\begin{align}
\label{con_hor4}
F_s=-\frac{2}{\pi} \; tan^{-1}(\frac{\delta \dot x}{\lambda})\; \mu F_N
\end{align}

The autors of \cite{juhasz2011beyond} adopted a model based on the dynamic and static friction coefficients ($\mu_{dyn}$ and $\mu_{stat}$) given by
\begin{align}
\label{con_hor5}
  F_s=\begin{cases}
             - sign(\delta\dot  x) \; \mu_{dyn} F_N \quad &, \quad \delta \dot x \leq v_{st}\\
            -\frac{\delta \dot x}{v_{st}} \; \mu_{stat} F_N \quad &, \quad \delta \dot x > v_{st}
            \end{cases}
\end{align}

In this model, $v_{st}$ is a threshold which distinguishes the static and dynamic friction phases. Jackson et al. 
\cite{jackson2016development} modified the Hollars friction model \cite{sherman2011simbody} to propose a complex frictional model to be continuous and differentiable.
\subsection{Proposed model}
\noindent
Due to the nature of the contact which depends on specifications of the interacting bodies, the variety of the proposed contact models is considerable (for example, see \cite{marques2017study} for a recent review in another field). Hence, comparing these contact models and selecting an appropriate one that is compatible with the reality for our problem is a demanding task. As a result, in order to model the contact between the robot and environment, we first specify the characteristics of a satisfactory contact model and then propose a model to comply with these requirements.

Expected characteristics of a contact model for the interaction between the feet of a biped robot and ground surface may be listed as:

\begin{itemize}
  \item  In the instances when the feet leave the ground surface or land on it, the contact force should be zero. This statement means that in these instances the ground surface should not exert any force to the contact points, because the deflection is zero. 

  \item The normal contact force should be unilateral during the interaction. As a matter of fact, due to the unilaterality premise of the contact between the feet and ground surface, the ground should not pull the feet.
  \item The contact elements should absorb some of the forces of the impact. This absorption depends on the material of the feet and ground.
  
  \item The maximum penetration depth should be adjustable. This value should be independent of the robot mass or velocity of the feet landing on the ground.
  
  \item Sensitivity of the model coefficients to the number of contact points and robot mass should be ignorable. 
  
  \item The friction model should have satisfactory resemblance to the reality. In fact, horizontal forces should depend on the normal forces. Furthermore, if the model is continuous, it can ease the numerical simulation. 
\end{itemize}

Based on these specifications, a linear model cannot satisfy these requirements. The reason is that the contact force in the landing instance is zero, if the velocity of the foot is zero. The proposed models in Eqs. (\ref{con_non1})-(\ref{con_non4}) are nonlinear models which satisfy the first 3 conditions  mentioned above; however they are sensitive to the impact conditions and the maximum penetration depth cannot be specified. 

Due to these shortcomings, we propose a nonlinear contact model to satisfy all of the above-mentioned factors. The effectiveness of the model that is proposed in this research has been verified through simulations of human walking on a treadmill \cite{dashkhaneh2014modeling}. The normal force in this model is stated as
\begin{align}
\label{con_prop}
F_N=-k_z \,  tan(\frac{\pi}{2 l_0} \delta z)-b_z |\delta z| \,\delta \dot z
\end{align}

In this equation $\delta z$ and $\delta \dot z$̇ are the deflection and rate of penetration, respectively. $k_z$ and $b_z$ are the stiffness and damping coefficients of the model. Also, $l_0$ is the maximum penetration depth. In this model, in the instances when the foot leaves the ground or lands on the ground surface ($\delta z=0$), the normal force is zero. Furthermore, by using appropriate coefficients, absorption of the impact is provided and the normal force is positive. Moreover, when $\delta z$ approaches $l_0$, the normal force approaches infinity. As a result, the maximum penetration depth can be adjusted. Finally, since in this model the stiffness and damping ratios are amplified by increasing penetration depth, sensitivity of the model to the impact conditions is negligible. 

In order to model the horizontal forces of contact, we use the pseudo-Coulomb model of Wojtyra \cite{wojtyra2003multibody}, namely
\begin{align}
\label{con_hor_prop}
F_s=-\frac{2}{\pi} \; tan^{-1}(\frac{\delta \dot x}{\lambda})\; \mu F_N
\end{align}

\begin{figure}
\centering
\includegraphics[clip,trim=1.5cm 18.5cm 2cm 3.5cm,width=12cm]{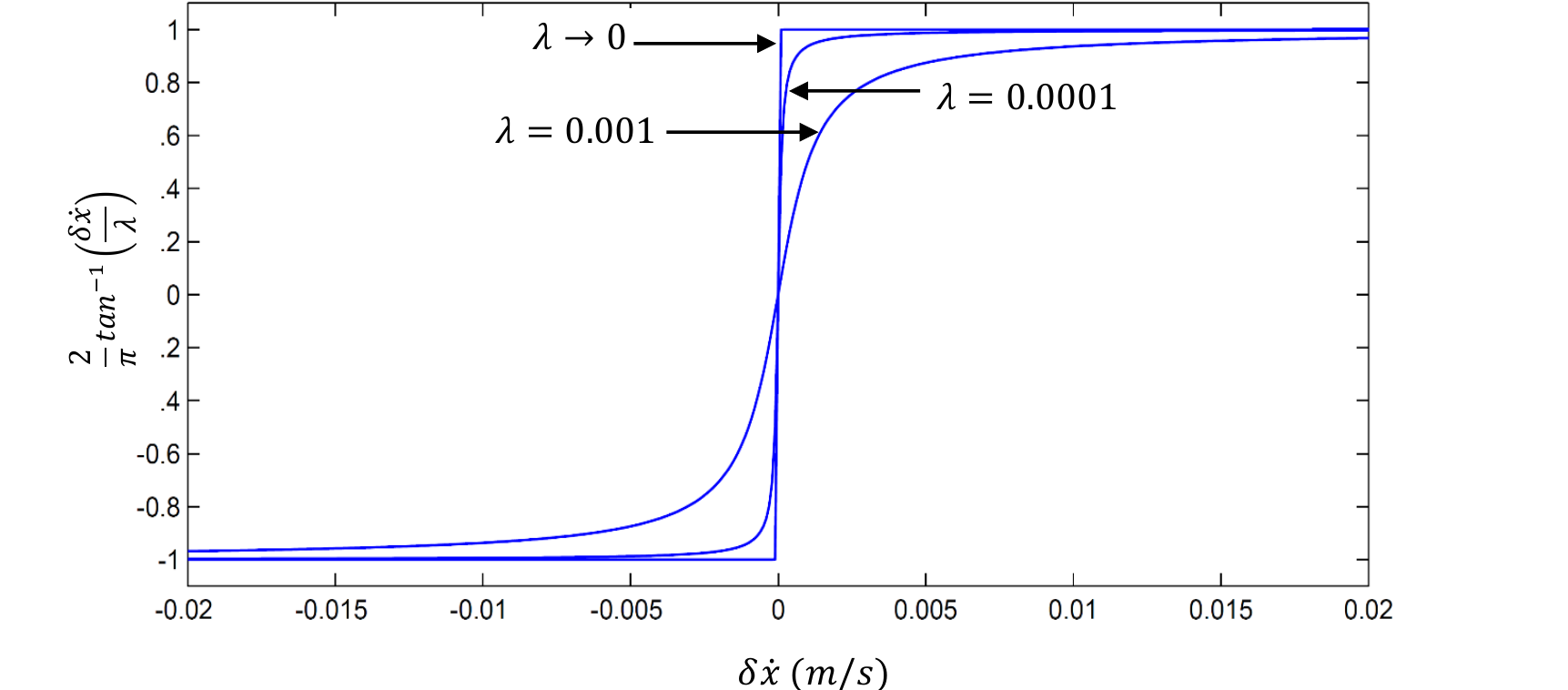}
\caption{Variation of the pseudo-Coulomb friction model}
\vspace{-1.5em}
\label{mu_lambda}
\end{figure}

In this equation, if $\lambda$ approaches zero, the horizontal force becomes:
\begin{align}
\label{con_hor_prop1}
F_s= \mu F_N \; sign (\delta \dot x)
\end{align}

which is the Coulomb model of friction (Fig. \ref{mu_lambda}). However, because the Coulomb model is discontinues in the vicinity of $\delta \dot x=0$, some oscillatory forces appear in numerical simulations. As a result, parameter $\lambda$ should be selected to make a compromise between the intimacy to the Coulomb friction model and generating non-oscillatory friction forces in numerical simulations. 

\subsection{Validation of the proposed model for normal contact}
\begin{figure}
\centering
\includegraphics[clip,trim=5cm 20.5cm 5cm 2.5cm,width=8cm]{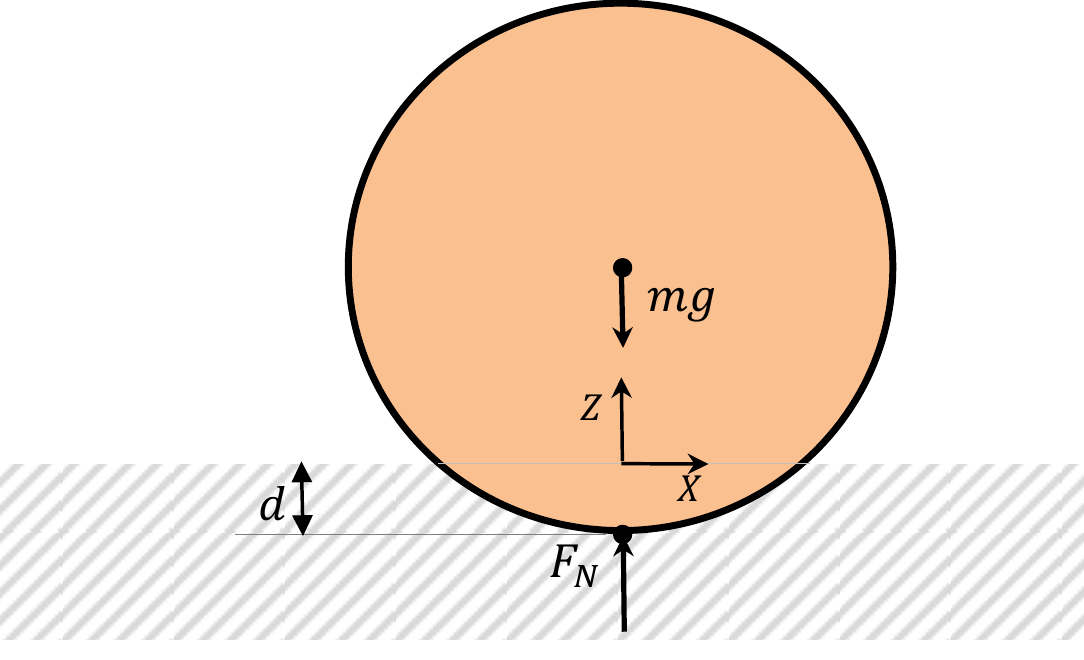}
\caption{A ball impacting the ground surface and corresponding parameters}
\vspace{-1.5em}
\label{ball}
\end{figure}

In order to show the effectiveness of the proposed contact model, we simulate the motion of  a ball with mass $m$ impacting the ground surface with speed $v$ and the corresponding penetration depth $d$ (Fig. \ref{ball}). We show that the proposed contact model is less sensitive to the mass of the body and impact velocity than those of the other models, which is a candidate for a model that passes our requirements, and also the penetration depth in our model can be limited. Based on our observations, these two points are more ambiguous compared to the other points, and we show them on a simple insightful example. Noteworthy is that here the shape of the object does not matter, as we have only one contact point. In the general case of a flat foot of the biped robot, we used several contact points, and we use the same equation for each contact point.\\
The inertial coordinate system is attached to the ground as depicted in Fig. \ref{ball}, while the normal reaction force is shown by $F_N$. As a result, we have the following conditions at the impact:
\begin{align}
\label{impact_par}
\delta z=-d \qquad , \qquad \delta \dot z=-v
\end{align}

\begin{figure}
\centering
\includegraphics[clip,trim=1.7cm 18.8cm 3.5cm 3.4cm,width=12cm]{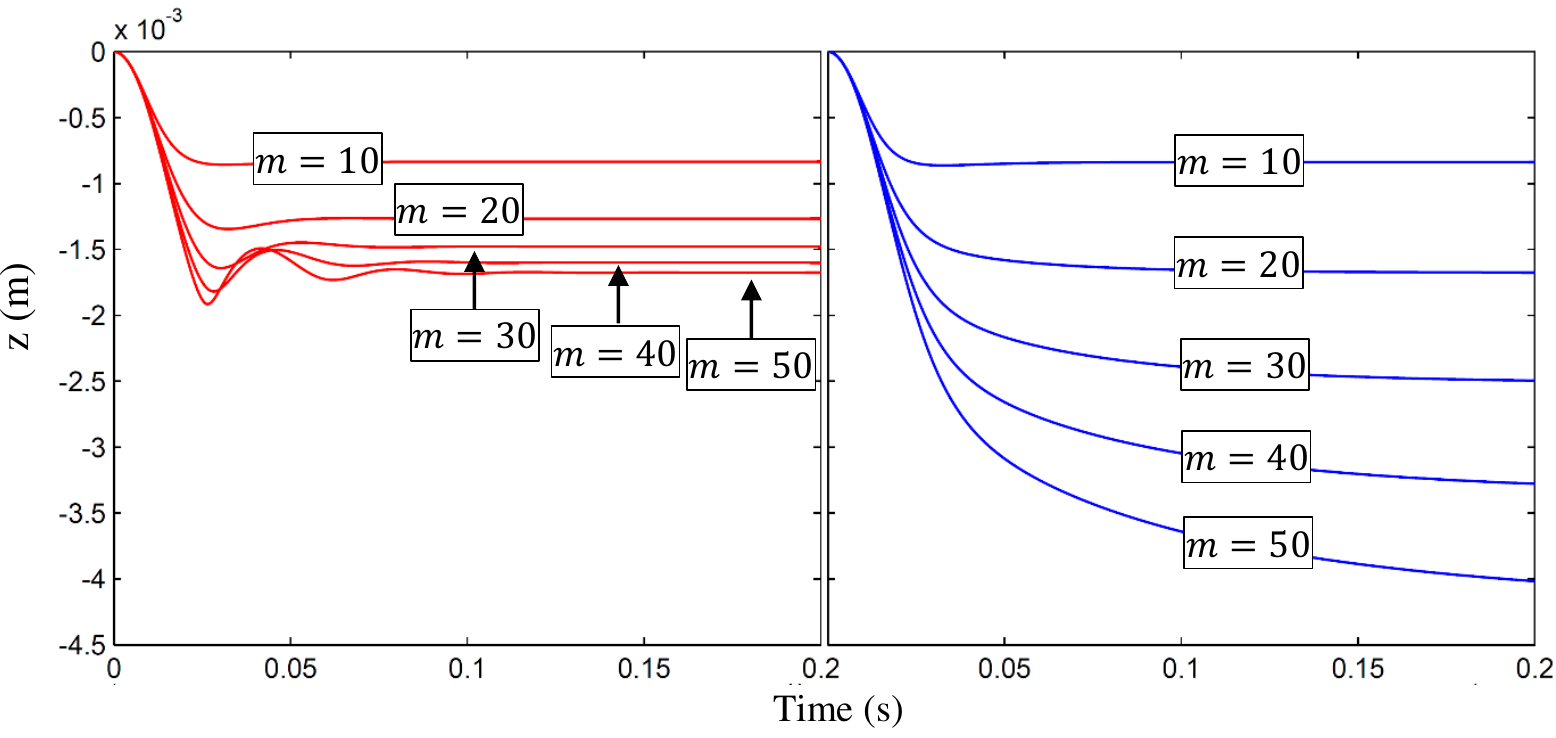}
\caption{Comparing the model in Eq. (\ref{con_non2}) (right) and the proposed model of Eq. (\ref{con_prop}) (left) where balls with different masses contact the ground surface with zero velocity}
\vspace{-1.5em}
\label{comp_penet}
\end{figure}

\begin{figure}
\centering
\includegraphics[clip,trim=1.7cm 18.8cm 3.5cm 2.5cm,width=12cm]{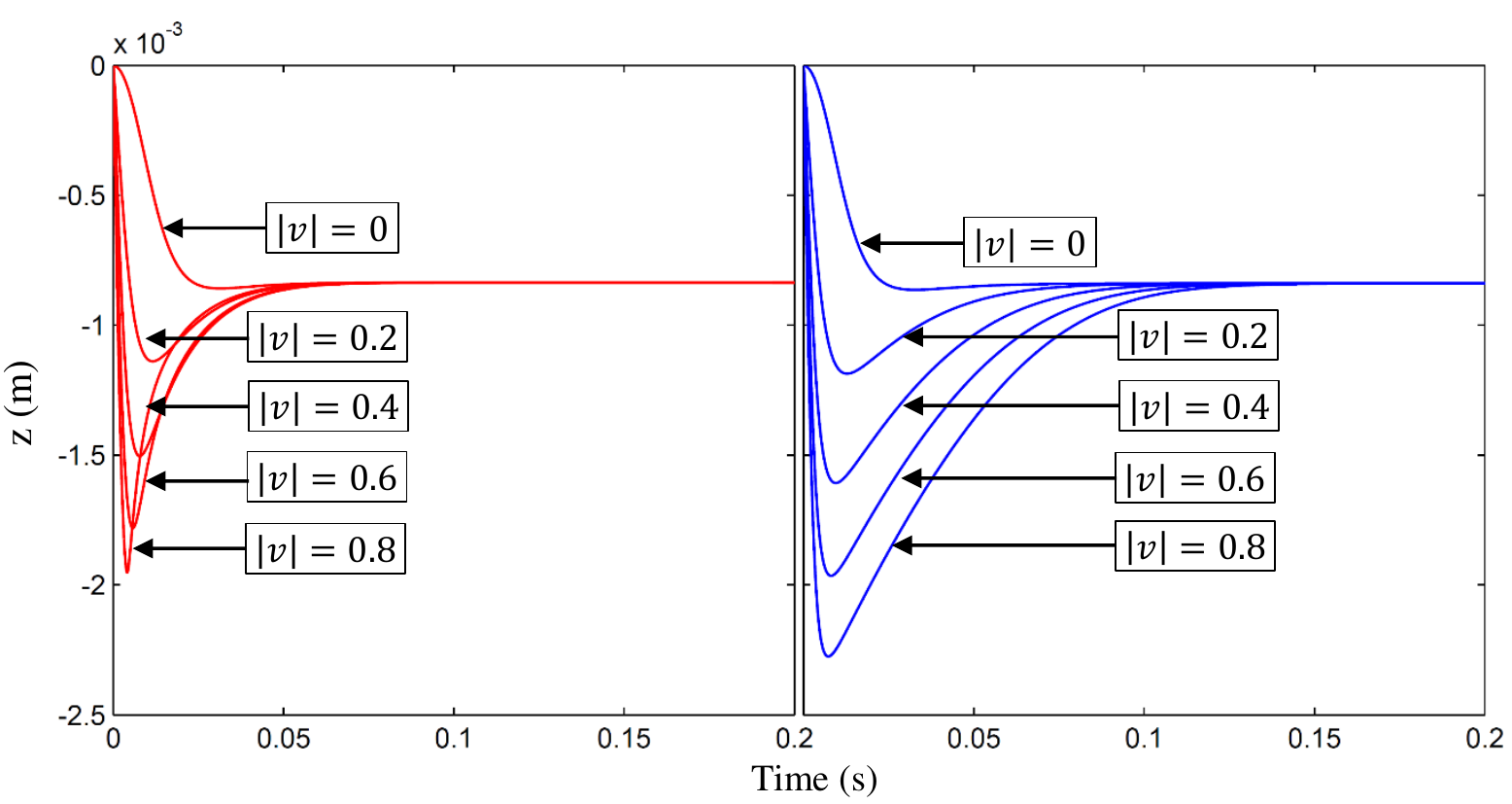}
\caption{Comparing the model in Eq. (\ref{con_non2}) (right) and the proposed model of Eq. (\ref{con_prop}) (left) where a ball with $m=10$ kg  contacts the ground surface with different velocities}
\vspace{-1.5em}
\label{comp_vel}
\end{figure}

We compare the proposed normal contact model Eq. (\ref{con_prop}) to the model specified in Eq. (\ref{con_non2}), which is a nonlinear viscoelastic model and a candidate for the contact model based on the specifications we mentioned. We selected the parameters of each model such that they yield similar behavior for simulating the motion of a ball with $m=10$ kg contacting the ground surface with zero velocity. Using the parameters in Table \ref{contact_properties}, both models show very close behavior as can be seen in Fig. \ref{comp_penet} with $m=10$ kg and in Fig. \ref{comp_vel} with zero impact velocity.

\begin{table}
\caption{The parameters used for contact modeling based on the model in Eq. (\ref{con_non2}) and the proposed model Eq. (\ref{con_prop})}
\label{contact_properties}       
\begin{tabular}{llll}
\hline\noalign{\smallskip}
Model in Eq. (\ref{con_non2}) &  & Proposed model of Eq. (\ref{con_prop}) &  \\
\noalign{\smallskip}\hline\noalign{\smallskip}
$k_z$ & 1.17e5 & $k_z$ & 1.0e5 \\
$b_z$ & 2.8e6 & $b_z$& 3.0e5 \\
  &   &$ l_0$& 0.002 \\
\noalign{\smallskip}\hline
\end{tabular}
\end{table}


Now, to compare these models, we examine two scenarios. In the first scenario, our goal is to investigate the sensitivity of the models to the variation of the ball mass. We can see in Fig. \ref{comp_penet} that varying the ball mass from 10 to 50 kg causes the change of the penetration depth from 0.8 to 4 mm for the model specified in Eq. (\ref{con_non2}). However, in our proposed model, the penetration depth is changed from 0.8 to 1.7 mm. This simulation shows that our proposed model is less sensitive to the change of the ball contacting the ground. The other point is that we cannot directly choose the maximum penetration depth in the model of Eq. (\ref{con_non2}), while the penetration depth obtained from our model is always less than $l_0$  (which is 0.2 mm in this simulation). Furthermore, as we can see in Fig. \ref{comp_penet}, the time required for our model to reach its steady state is way less than for the model of Eq. (\ref{con_non2}). As it can be observed in this figure, for the case of $m=10$ kg, after 0.025 s both models reach their steady state. However, increasing the ball mass to 50 kg, the settling time of the model in Eq. (\ref{con_non2}) increases to 0.2 s, while in our model the settling time remains below 0.06 s.

In the second scenario, we change the impact velocity and compare the response of the two models. The results for this case are shown in Fig. \ref{comp_vel}, where $m=10$ kg and the impact velocity changes from $\delta \dot z=-v=0$ to $\delta \dot z=-v=0.8$ m/s. As we can see in this figure, increasing the impact velocity from 0.8 to 2.2 m/s causes an increase from 0.8 to 2.2 mm in the penetration depth in the model of Eq. (\ref{con_non2}). However, in our model the penetration depth varies from 0.8 to 1.9 mm, which is less than the maximum penetration depth $l_0=0.2$ mm. Furthermore, we can see that by increasing the impact velocity the settling time of the model in Eq. (\ref{con_non2}) increases from 0.025 to 0.12 s, while in our model the settling time is less than 0.06 s. 

Based on these scenarios, we can see that our model for the contact in the normal direction matches the requirements for an ideal contact model.

\section{Results and discussion}
\label{simulation}
\begin{figure}
\centering
\includegraphics[clip,trim=4cm 17cm 4cm 3.5cm,width=12cm]{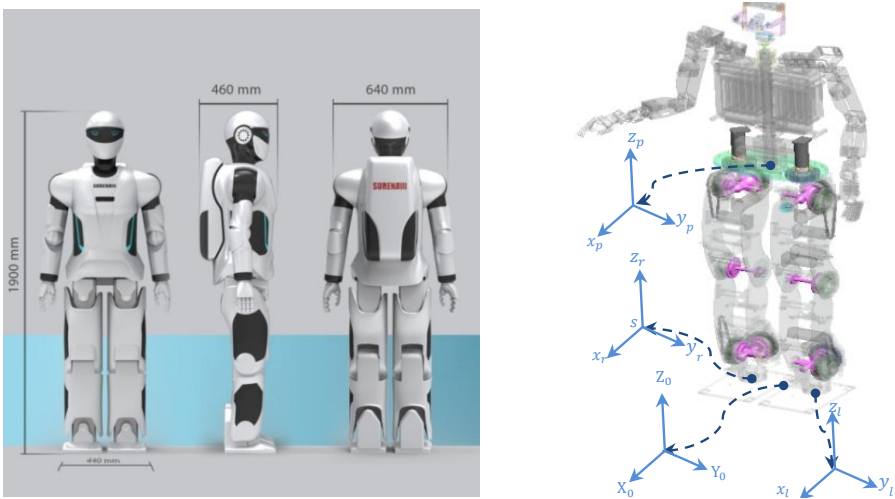}
\caption{SURENA III, a humanoid robot designed and fabricated in CAST, University of Tehran.}
\vspace{-1.5em}
\label{surena}
\end{figure}

In this section we compare the models developed with rigid and compliant contact to the empirical result obtained from implementing a walking pattern on the humanoid robot Surena III (Fig. \ref{surena}). This robot is composed of 12 DOF in its lower-body and 19 DOF in its upper-body. This robot has 6 DOF in each leg ( 3 DOF for hip, 1 DOF for knee, 2 DOF for ankle), 7 DOF in each arm ( 3 DOF for shoulder, 1 DOF for elbow, 3 DOF for wrist), one DOF in each hand (a simple gripper), one DOF in the torso and 2 DOF in the neck (Fig. \ref{surena}, right). Actuation of the lower-body is done by EC motors in each joint. The power transmission system is comprised of belts and pulleys, and harmonic drives. For the upper-body, servomotors with embedded gearbox and driver are employed. The sensory layout includes incremental and absolute encoders on the motor output and gearbox output of each joint, 6-axis force/torque sensors embedded to the ankles, and an IMU mounted on the upper-body. The mass properties and geometric attributes of this robot are specified in Table \ref{properties}.

\begin{table}
\caption{Mass properties and geometric attributes of the SURENA III humanoid robot}
\label{properties}       
\begin{tabular}{llll}
\hline\noalign{\smallskip}
Link & Mass (gr) & Parameter & Value (mm)  \\
\noalign{\smallskip}\hline\noalign{\smallskip}
Foot & 3859 & Foot length & 265 \\
Ankle & 2236 & Foot width & 160 \\
Shank & 4561 & Ankle joints height & 98 \\
Thigh &6327 & Shank length & 360 \\
Pelvis & 17800 & Thigh length & 360 \\
Upper-body & 36234 & Distance between hip-rotation joints & 230 \\
  &   & Distance between hip and pelvis & 115 \\
    &   & Distance between pelvis and head & 967 \\
    \noalign{\smallskip}\hline\noalign{\smallskip}
    Total Weight & 88 (kg) & Total Height & 1.90 (m) \\
\noalign{\smallskip}\hline
\end{tabular}
\end{table}

To model the robot dynamics with rigid contact, we use the procedure explained in Sect.~\ref{rigid}. We compute the inverse dynamics for both the SSP and DSP walking phases (Eqs. (\ref{dynamics_ssp_sol}) and (\ref{dynamics_dsp_psol})). We use the Yobotics physics engine \cite{pratt2000exploiting} in order to simulate motion of the robot with the proposed compliant contact model (Eqs.(\ref{con_prop}) and (\ref{con_hor_prop})). The lower-body of the robot is modeled using these physical software tools, while the upper-body is modeled by a single rigid body (Fig. \ref{walking}). In the rest of this section our focus will be on bipedal walking of the robot, while we demonstrated simulation results for the whole body motion with the upper-body joints in our earlier work \cite{khadiv2014dynamics}. In order to increase the precision of the multibody model in estimating joint torques, we identify the drive system dynamics (see Appendix A) and add it to the multibody dynamics with rigid and compliant contact models (see Fig. \ref{block_diagram}):
\begin{align}
\label{tau_final}
\tau=\tau_{multibody}+\tau_{drive}
\end{align}

In order to compare the results obtained from the developed models and experiment, we generate a walking pattern with the speed of 0.5 km/h \cite{khadiv2015optimal} and apply the joint trajectories to the models (with rigid and compliant contact) and experimental setup. The joint torques for the models are obtained using Eq. (\ref{tau_final}). For the real robot, we measured the motors' current and logged it during the motion. Then, we computed joint torques by multiplying the measured current to the motor torque constant. 

\begin{figure}
\centering
\includegraphics[clip,trim=2.5cm 16.cm 1cm 2.5cm,width=12cm]{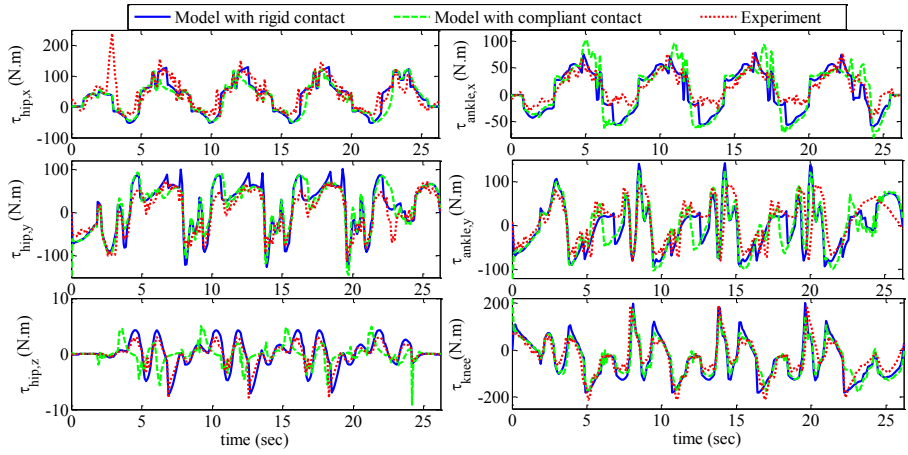}
\caption{Time history of the joint torques during three gait cycles}
\vspace{-1.5em}
\label{torque}
\end{figure}

Figure \ref{torque} compares the obtained results from the models and experiment for the lower-body joints. As we can observe in this figure, both models fairly estimate the required torque for all the joints. This can be highlighted, when we can see that both models estimate the maximum and minimum values of the experimental data precisely, while they also follow the trend of the experimental torque profile. However, at some parts of the torque profiles, we can see some error between the models and experimental results. This stems from many factors such as error in drive system identification, difference between the contact model and real contact, error in parameters of the robot, etc. Furthermore, we can see some outliers in the experimental measurements such as of the hip in thee $x$-direction at $t=2.5$ s. The other point we can see in this figure is that in the torque profile of the ankle joint in the $x$-direction, there exists an offset between the models and experimental measurements for the negative values. This observation suggests that the robot sways in the simulation more than the real experiment. 

 Based on the obtained results in Fig. \ref{torque}, it is hard to say which model better estimates the joint torques. In fact, based on Eq. (\ref{tau_final}), the required torque at each joint depends on both the drive system and multibody dynamics, while the contact dynamics just affects the multibody dynamics. Since the gear ratio at each joint of our robot is relatively high (360 or 475), the drive system dynamics is the dominant part of the torque required at each joint. Moreover, further inspection on the equations of motion Eq. (\ref{dynamics}), together with the corresponding generalized forces Eqs. (\ref{gen_force_ssp}) and (\ref{gen_force_dsp}) reveals that the required torque of the multibody dynamics is a sum of both the projection of the contact forces to the generalized coordinates space and the inertial and gravitational effects. However, the gravitational and inertial effects do not directly depend on the rigid or compliant nature of the contact. Due to these facts, we decided to avoid quantitative error analysis because we could end up presenting very vague interpretation of the error analysis. We can then conclude that both models yield satisfactory estimation of the required joint torque at each joint, while employing the model with rigid contact for computing the joint torques is preferable due to its less computation burden. As a result, for optimizing walking patterns in terms of torque-based cost functions, analyzing generated gaits and designing model-based controllers the model with rigid contact can be effectively used.
 
\begin{figure}
\centering
\includegraphics[clip,trim=2.5cm 16.cm 1cm 2.5cm,width=10cm]{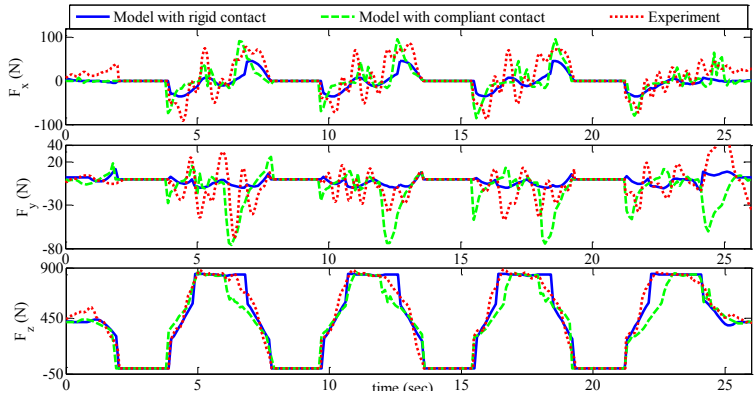}
\caption{Time history of left foot-ground interacting forces during three gait cycles}
\vspace{-1.5em}
\label{forces}
\end{figure}

In Fig. \ref{forces}, we compared the interaction forces obtained from the model with rigid contact and the model with compliant contact to the experimental results measured using the 6-axis force/torque sensors attached to the ankles. As it can be observed, the model with compliant contact (green) better estimates the horizontal interaction forces compared to the model with rigid contact (blue). In fact, the model with compliant contact better estimates the maximum and minimum values which have a significant role in estimating the foot slippage. As a result, the proposed compliant contact model yields more realistic results and it can predict slippage of the feet better than the model with rigid contact. The reason is that in reality there always exists some compliance between the robot feet and ground surface.

 For the normal component of the interaction forces, both models yield satisfactory results. Furthermore, it can be seen in this figure that the normal force is always positive and the unilaterality constraint is respected in both models. The other advantage of the model with compliant contact compared to the model with rigid contact is that it is not required to specify the walking phases in advance. In fact, in this model, if the swing foot lands on the ground sooner or later than it is expected \cite{khadiv2017online}, or maybe it lands on the edge of the foot, these effects are simulated and consistent results are computed. However, in the model with rigid contact, the gait phases should be specified and the solution for each phase should be computed separately. Hence, if the swing foot lands on the ground with non-zero velocity (due to uncertainties), we need to take into account the impact phenomenon separately, while for the model with compliant contact it is simulated automatically. Hence, we can conclude that in order to simulate the generated gaits or designed controllers, the model with compliant contact can yield more realistic results even in the presence of uncertainties and unwanted disturbances such as impact.
 
\begin{figure}
\centering
\includegraphics[clip,trim=2.5cm 16.cm 1cm 2.5cm,width=10cm]{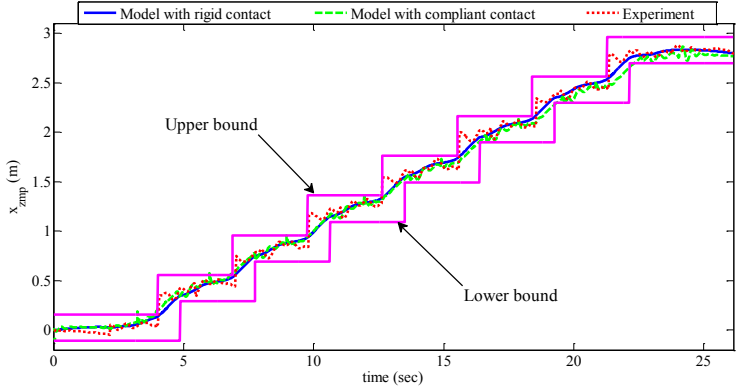}
\caption{Time history of the ZMP in sagittal direction during three gait cycles}
\vspace{-1.5em}
\label{zmp_x}
\end{figure}

\begin{figure}
\centering
\includegraphics[clip,trim=2.5cm 16.cm 1cm 2.5cm,width=10cm]{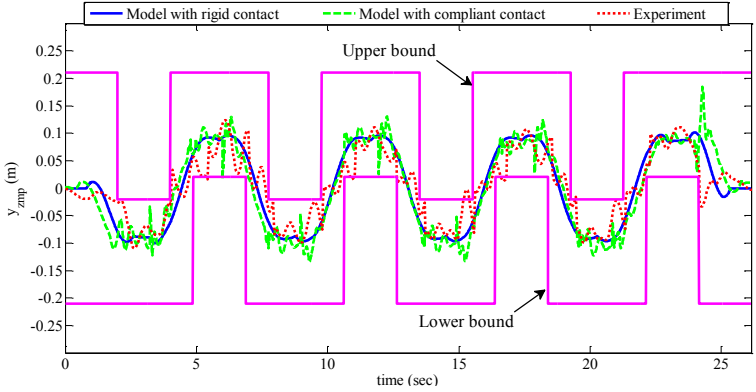}
\caption{Time history of the ZMP in lateral direction during three gait cycles}
\vspace{-1.5em}
\label{zmp_y}
\end{figure}

\begin{figure}
\centering
\includegraphics[clip,trim=2.5cm 16.cm 1cm 2.5cm,width=14cm]{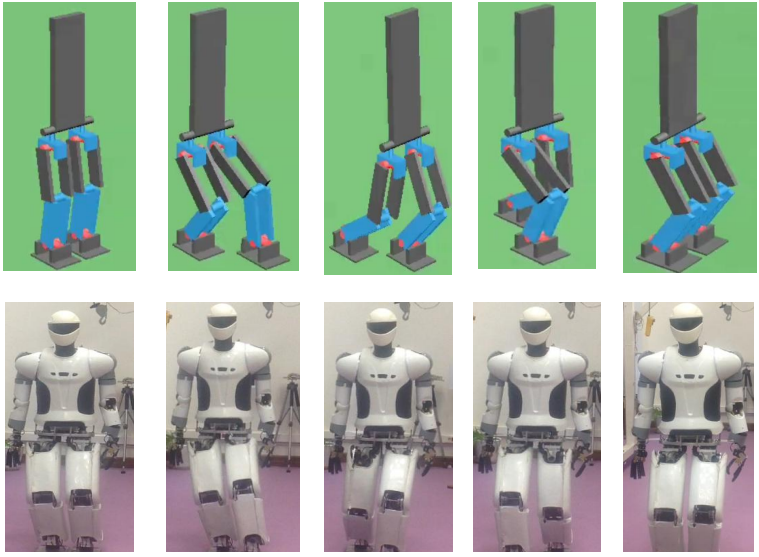}
\caption{Simulation and experimental validation of walking at speed 0.5 km/h. ( SURENA III humanoid robot which is designed and fabricated in CAST)}
\vspace{-1.5em}
\label{walking}
\end{figure}

As we discussed in Sect.~\ref{rigid}, the proposed inverse dynamics solutions are valid as long as the feasibility constraints are satisfied. The first constraint, i. e. the normal forces being positive and unilateral, has been shown to be valid in Fig. \ref{forces}. For slippage avoidance, the required friction coefficient should be less than the available friction coefficient between the feet and surface. As we observed in Fig. \ref{forces}, the model with compliant contact can better estimate the required friction coefficient for a given walking pattern. To check that the stance foot (feet) does not rotate around the edges of the support polygon (as it is planned \cite{khadiv2015optimal}), in Figs. \ref{zmp_x} and \ref{zmp_y} the ZMP trajectories obtained from the models with compliant contact and rigid contact are shown. As it can be observed, the ZMP obtained from the model with rigid contact is strictly inside the support polygon which guarantees that the stance foot (feet) does not rotate. Furthermore, we can see that the fluctuations of the ZMP trajectory for the model with compliant contact are more than the model with rigid contact. This is because of the fact that deflections of the contact points generate more oscillatory interaction moments. This is a more realistic situation compared to the model with rigid contact. Hence, the generated gaits that are feasible based on simulation of the model with compliant contact are more reliable to be tested on the real robot.

As we discussed in Sect.~\ref{rigid}, if the feasibility constraints are not satisfied, and the stance foot starts to rotate due to the uncertainties, then the obtained solutions are not valid for the model with rigid contact and consistent solution with the real scenario should be computed. This means that for different contact conditions, different solutions should be computed. For instance, we computed the inverse dynamics solution for various gait phases such as toe-off and heel-contact in our previous works \cite{ezati2015effects,sadedel2016adding,sadedel2016investigation}. However, the model with compliant contact can cope with this situation and the results are automatically adapted to the real contact condition.
Finally, Fig. 12 illustrates snapshots of the robot walking in the simulation environment (Yobotics) and the real environment for one gait cycle. 

\section{Conclusion}
\label{conclusion}
In this paper, we have compared two dominant contact modeling approaches in the field of bipedal locomotion, i. e., rigid and compliant contact. For the rigid contact case, we modeled the multibody of the robot consistent with holonomic constraints, and computed the inverse dynamics solution in each phase. For the compliant contact case, we modeled the multibody of the robot in a physics engine, while we used our  proposed nonlinear contact model. In order to conduct the comparison, we analyzed the results obtained from applying a feasible walking pattern to both models and the experimental setup. This comparison revealed that both models yield satisfactory joint torques estimation. Since the model with rigid contact is less complex compared to the model with compliant contact, this model can be effectively used in optimizing and analyzing walking patterns, as well as designing model-based walking controllers. However, this model does not take into account the impact phenomenon, the walking phases should be specified in advance for this model, and the computed interaction forces and moments are not precise enough for simulating the robot motion. The model with compliant contact without having these constraints but with more complexity which demands more computational cost is suitable for simulating motion of the robot in a more realistic scenario.

\begin{acknowledgements}
The authors would like to express deep gratitude to the Industrial Development
and Renovation Organization of Iran (IDRO) and Iran National Science Foundation (INSF) for their
financial support (Project Number: 95849278) to develop the SURENA III humanoid robot. We
further thank to the members of CAST for their valuable participation in the design and fabrication
of the robot.
\end{acknowledgements}

\section*{Appendix A : Drive system identification}
\label{identification}
\subsection*{A.1  Method}
Our main goal in this section is to employ a simple representation which can replicate drive system dynamics behavior to a desired level of accuracy. The major effects that may be taken into account for identification of the drive system are the effective inertia, Coulomb friction, and viscous friction of the system and some other load-dependent terms \cite{taghirad1998modeling}. Hence, the considered general model may be represented as
\begin{align}
\label{tau_drive}
\tau_{drive}=j\ddot \theta+b \dot \theta+ c \dot \theta^3+f \; sign(\dot \theta) + ...
\end{align}

in which $j$ is the inertia parameter, $b$, and $c$ are the parameters for viscous friction (and potentially the electromotive force of the motor ,back emf), and $f$ is the Coulomb friction parameter . Furthermore, $\tau_{drive}$ is the torque that is exerted by the motor, and $\theta$ is the joint angle which is measured by the encoder mounted at the output of the drive system.

In order to identify unknown parameters, a least squares curve fitting approach is adopted. The input trajectories that are considered for the identification procedure are the joints trajectories for various walking speeds. These trajectories include high and low velocity, as well as low and high frequency commands. Moreover, the motor torque can be computed by measuring the motor current and multiplying it to the motor torque constant, or exploiting a torque sensor at the output of the drive system. As a result, the linear regression model may be specified as
\begin{align}
\label{regression}
\tau_{m \times 1}=\begin{bmatrix} \ddot \theta & \dot \theta & \dot \theta^3 & sign(\dot \theta) \end{bmatrix}_{m \times n} \begin{bmatrix}  j \\ b \\c \\ f \\ \vdots \end{bmatrix}_{n \times 1}
\end{align}

in which $m$ specifies samples that are taken from the measured values during one experiment, and $n$ is the number of parameters that should be identified. It should be noted that a necessary condition for identification is that $m$ should be greater than $n$. Using the Moore-Penrose inverse (left pseudo-inverse), the identification routine is carried out to minimize the quadratic norm of the parameters error
\begin{align}
\label{pinv_ident}
\begin{bmatrix}  j \\ b \\c \\ f \\ \vdots \end{bmatrix}_{n \times 1}=\begin{bmatrix} \ddot \theta & \dot \theta & \dot \theta^3 & sign(\dot \theta) \end{bmatrix}^\dagger_{n \times m}  \; \tau_{m \times 1}
\end{align}

Using this method, for each experiment, a set of parameters may be obtained. As a result, in order to obtain a model which is valid for a wide range of experiments, the average value of obtained parameters may be considered as a candidate for the overall model. The obtained identified model will be acceptable, provided that it is valid for a wide range of experiments. This consistency can be evaluated using the consistency measure \cite{taghirad1998modeling}, which is the ratio of the standard deviation $STDV$ to the average value $AVG$ of each parameter, namely 
\begin{align}
\label{con_mea}
C.M. = \frac{STDV}{AVG}
\end{align}

If the consistency measures obtained for all parameters are in a desired range \cite{taghirad1998modeling}, the obtained model is acceptable. Otherwise, some other terms should be added to the model to improve the consistency between the obtained parameters for various experiments. 

\subsection*{A.2  Results}
\begin{figure}
\centering
\includegraphics[clip,trim=2.5cm 20cm 4cm 3cm,width=10cm]{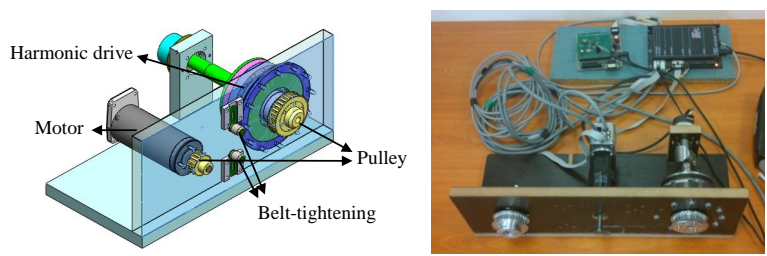}
\caption{The developed test-bed for system identification, system components (left), experimental setup (right)}
\vspace{-1.5em}
\label{drive}
\end{figure}

The drive system of the SURENA III humanoid robot is composed of EC motors, pulleys and timing belts, and harmonic drive gears. In Fig. \ref{drive}, the components of the drive system and the the developed test-stand for the identification purpose are shown. The three major effects that are taken into account in our identification routine are the effective inertia, Coulomb friction, and viscous friction of the system. Hence, the considered model may be represented as
\begin{align}
\label{motor}
\tau_{drive}=N_p N_h k_m i=j\ddot \theta+b \dot \theta+f \; sign(\dot \theta)
\end{align}

in which $N_p$ and $N_h$ are the pulley and harmonic reduction ratios, $k_m$ is the motor torque constant, and $i$ is the motor input current; $j$ , $b$ , and $f$ are the parameters that should be identified. It should be noted that since no torque sensor is available at the output of the harmonic drive, the load-dependent terms are not included in this model.

Using the procedure that has been described in Sect. A.1, the identification routine is carried out and the obtained values for the identified model are summarized in Table \ref{ident_result}. These values are obtained by applying 5 experiments using the knee joint motion for walking from 0.3 to 0.7 km/h. 

\begin{table}
\caption{Obtained results from identification of the drive system, $j$ is the estimated inertia, $b$ is the estimated viscus friction coefficient, and $f$ is the estimated Coulomb friction coefficient.}
\label{ident_result}       
\begin{tabular}{llll}
\hline\noalign{\smallskip}
Experiment No. & $j$  & $b$ & $f$ \\
\noalign{\smallskip}\hline\noalign{\smallskip}
1 & 10.51 & 116.48& 24.34 \\
2 & 9.84 & 105.00& 25.25 \\
3 & 1.37 & 88.11& 26.34 \\
4 & 13.017 & 58.32& 24.20 \\
5 & 5.96 & 68.77& 24.04 \\
\noalign{\smallskip}\hline\noalign{\smallskip}
AVG & 8.14 & 87.34& 24.83 \\
STDV & 2.07 & 21.67& 0.86 \\
C. M. (\%)& 25.43 & 24.8& 3.47 \\
\noalign{\smallskip}\hline
\end{tabular}
\end{table}

\begin{figure}
\centering
\includegraphics[clip,trim=2.5cm 16.6cm 2cm 2.5cm,width=10cm]{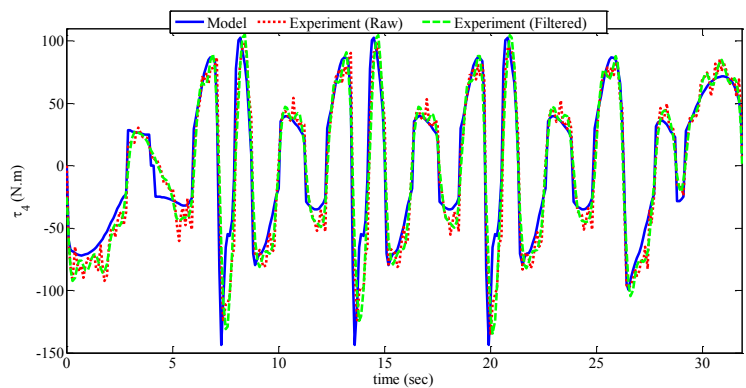}
\caption{The obtained results for the knee joint identification}
\vspace{-1.5em}
\label{drive_ident}
\end{figure}

\begin{figure}
\centering
\includegraphics[clip,trim=2.5cm 16.4cm 2cm 2.5cm,width=10cm]{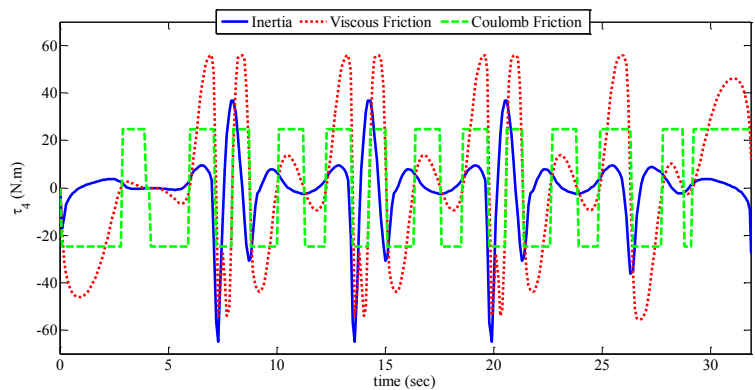}
\caption{Inertia, viscous and coulomb friction effects on the knee joint torque}
\vspace{-1.5em}
\label{ident_portion}
\end{figure}

As it can be observed in Table \ref{ident_result}, the obtained consistency measure for the parameter $f$ is absolutely acceptable. Also, for the parameters $j$ and $b$ this measure is satisfactory for our comparison purposes \cite{taghirad1998modeling}. Therefore, the obtained model with average values moderately estimates the dynamics of the drive system for various walking speeds. 

In Fig. \ref{drive_ident}, the identified model for the drive system of the knee joint is plotted. As it can be observed, the model fairly estimates the behavior of the drive system at the speed of 0.5 km/h. Also, in order to analyze effects of the components of the model, in Fig. \ref{ident_portion} each effect is plotted separately. As it can be seen in this figure, the inertia and viscous friction have a dominant effect in high velocity and acceleration motions. However, the Coulomb friction effect has an approximately constant value which varies when the direction of motion changes.

\bibliography{Master}
\bibliographystyle{spmpsci}      

\end{document}